\documentclass{bmcart}

\usepackage{comment}
\usepackage{fancyvrb}
\usepackage{graphicx}
\usepackage{epsfig}
\usepackage[ragged]{footmisc}
\usepackage{ragged2e}
\usepackage{hyperref}
\usepackage{color}
\usepackage{bm}
\usepackage{gensymb}
\usepackage[utf8]{inputenc} 

\startlocaldefs

\definecolor{blue}{rgb}{0,0,1}
\definecolor{red}{rgb}{1,0,0}

\endlocaldefs

\begin{document}

\begin{frontmatter}

\begin{fmbox}
\dochead{Research}

\title{Application of the SP Theory of Intelligence to the Understanding of Natural Vision and the Development of Computer Vision}


\author[
   addressref={aff1},                   
   email={jgw@cognitionresearch.org}    
]{\inits{JG}\fnm{J Gerard} \snm{Wolff}}

\address[id=aff1]{
  \orgname{CognitionResearch.org}, 
  \city{Menai Bridge},                        
  \cny{UK}                                    
}

\end{fmbox}


\begin{abstractbox}

\begin{abstract}

The {\em SP theory of intelligence} aims to simplify and integrate concepts in computing and cognition, with information compression as a unifying theme. This article is about how the SP theory may, with advantage, be applied to the understanding of natural vision and the development of computer vision. Potential benefits include an overall simplification of concepts in a universal framework for knowledge and seamless integration of vision with other sensory modalities and other aspects of intelligence. Low level perceptual features such as edges or corners may be identified by the extraction of redundancy in uniform areas in the manner of the run-length encoding technique for information compression. The concept of {\em multiple alignment} in the SP theory may be applied to the recognition of objects, and to scene analysis, with a hierarchy of parts and sub-parts, at multiple levels of abstraction, and with family-resemblance or polythetic categories. The theory has potential for the unsupervised learning of visual objects and classes of objects, and suggests how coherent concepts may be derived from fragments. As in natural vision, both recognition and learning in the SP system are robust in the face of errors of omission, commission and substitution. The theory suggests how, via vision, we may piece together a knowledge of the three-dimensional structure of objects and of our environment, it provides an account of how we may see things that are not objectively present in an image, how we may recognise something despite variations in the size of its retinal image, and how raster graphics and vector graphics may be unified. And it has things to say about the phenomena of lightness constancy and colour constancy, the role of context in recognition, ambiguities in visual perception, and the integration of vision with other senses and other aspects of intelligence.

\end{abstract}



\end{abstractbox}

\end{frontmatter}

\section{Introduction}

The {\em SP theory of intelligence}, introduced below, aims to simplify and integrate ideas across artificial intelligence, mainstream computing, and human perception and cognition, with information compression as a unifying theme. This article is about how the SP theory may, with advantage, be applied to the understanding of natural vision and the development of computer vision, and to discuss associated issues.

In these areas, potential benefits of the SP theory and its broad perspective include:

\begin{itemize}

 \item Developing concepts that combine {\em simplicity} with descriptive or explanatory {\em power}, in accordance with Occam's Razor \cite[Sections 2 and 4]{sp_benefits_apps}.

 \item Deeper insights and better solutions to problems \cite[Section 6]{sp_benefits_apps}.

 \item With natural vision, developing concepts that are consistent with evidence that general principles apply across different parts of the brain and across different aspects of brain function \cite[Section III-A.7]{sp_big_data}.

 \item The seamless integration of artificial vision with other sensory modalities, and with other aspects of intelligence such unsupervised learning, pattern recognition, reasoning, planning, problem solving, and more \cite[Section 7]{sp_benefits_apps}---with consequent benefits in terms of the versatility and adaptability of artificial systems.

 \item Contributing to the development of a {\em universal framework for the representation and processing of diverse kinds of knowledge} (UFK) and thus helping to overcome the problem of variety in big data \cite[Section III]{sp_big_data}.

\end{itemize}

The central idea is that, in accordance long-established principles in science, we should aim for theories with broad scope and avoid micro-theories that only work in restricted areas. If one's view is too narrow, it may be difficult to grasp the big picture---like the blind men trying to understand an elephant. 

The next section describes the SP theory in outline, and subsequent sections discuss how it may be applied to the understanding of natural vision and the development of artificial vision. These two aspects of vision are discussed together throughout the article, since each one may illuminate the other.

\section{Outline of the SP theory}\label{sp_outline_section}

\sloppy The SP theory is described fairly fully in an extended overview \cite{sp_extended_overview}, with enough detail to ensure that the rest of this article makes sense. The most comprehensive description of the theory and its applications is in the book, {\em Unifying Computing and Cognition} \cite{wolff_2006}. Applications of the theory are described in \cite{sp_big_data,sp_benefits_apps,wolff_sp_intelligent_database,wolff_medical_diagnosis} and other articles, details of which may be found via \href{http://www.cognitionresearch.org/sp.htm}{www.cognitionresearch.org/sp.htm}.

\begin{figure}[!htbp]
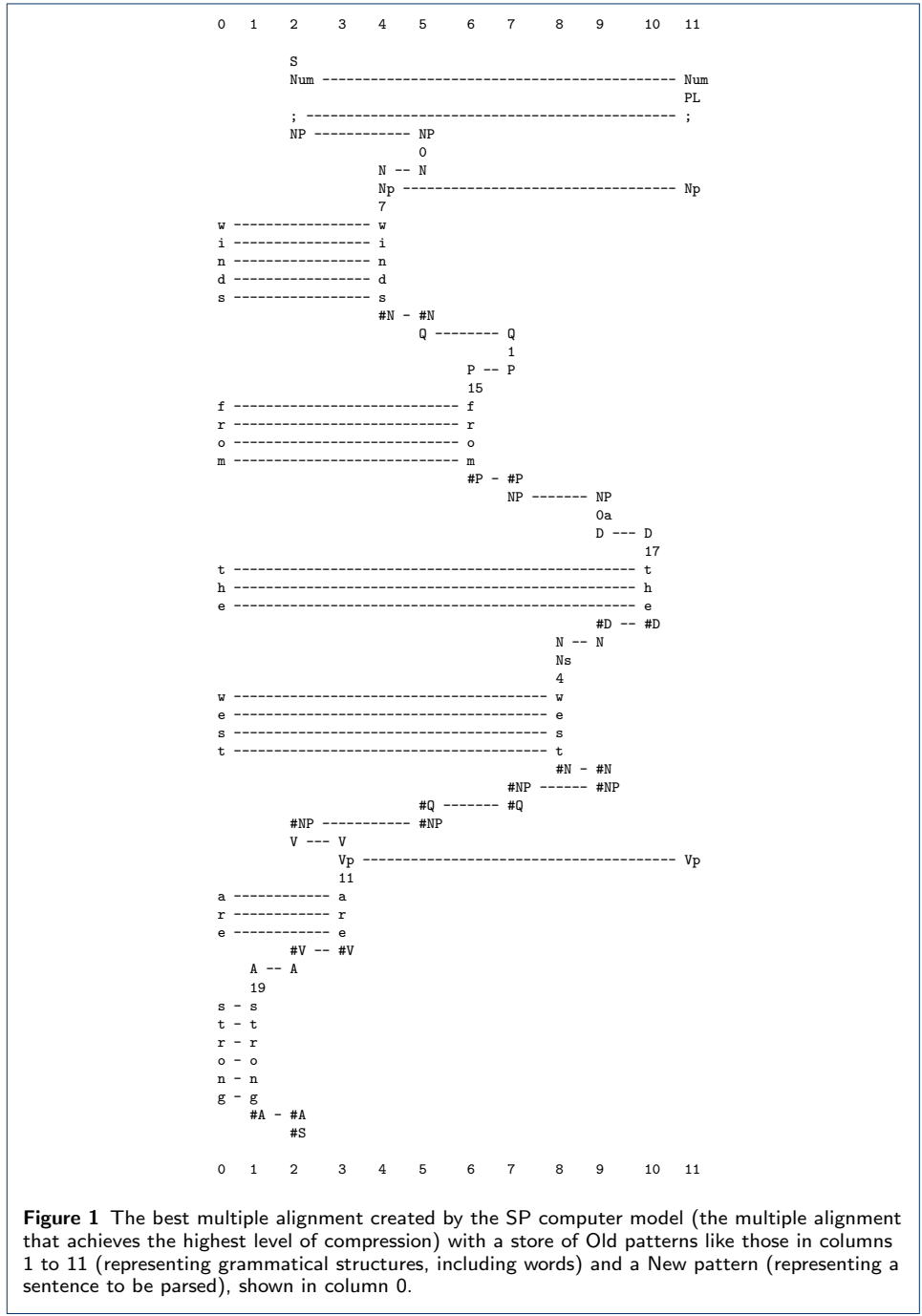

\fontsize{06.00pt}{07.20pt}
\centering
{\bf
\begin{BVerbatim}
0   1    2     3    4    5     6    7     8    9     10   11

         S
         Num -------------------------------------------- Num
                                                          PL
         ; ---------------------------------------------- ;
         NP ------------ NP
                         0
                    N -- N
                    Np ---------------------------------- Np
                    7
w ----------------- w
i ----------------- i
n ----------------- n
d ----------------- d
s ----------------- s
                    #N - #N
                         Q -------- Q
                                    1
                               P -- P
                               15
f ---------------------------- f
r ---------------------------- r
o ---------------------------- o
m ---------------------------- m
                               #P - #P
                                    NP ------- NP
                                               0a
                                               D --- D
                                                     17
t -------------------------------------------------- t
h -------------------------------------------------- h
e -------------------------------------------------- e
                                               #D -- #D
                                          N -- N
                                          Ns
                                          4
w --------------------------------------- w
e --------------------------------------- e
s --------------------------------------- s
t --------------------------------------- t
                                          #N - #N
                                    #NP ------ #NP
                         #Q ------- #Q
         #NP ----------- #NP
         V --- V
               Vp --------------------------------------- Vp
               11
a ------------ a
r ------------ r
e ------------ e
         #V -- #V
    A -- A
    19
s - s
t - t
r - r
o - o
n - n
g - g
    #A - #A
         #S

0   1    2     3    4    5     6    7     8    9     10   11
\end{BVerbatim}
}
\caption{The best multiple alignment created by the SP computer model (the multiple alignment that achieves the highest level of compression) with a store of Old patterns like those in columns 1 to 11 (representing grammatical structures, including words) and a New pattern (representing a sentence to be parsed), shown in column 0.}
\label{winds_from_west_figure}
\end{figure}

In broad terms, the SP theory has three main elements:

\begin{itemize}

\item All kinds of knowledge are represented with {\em patterns}: arrays of atomic symbols in one or two dimensions.

\item At the heart of the system is compression of information via the matching and unification (merging) of patterns, and the building of {\em multiple alignments} like the one shown in Figure \ref{winds_from_west_figure}.

\item The system learns by compressing {\em New} patterns to create {\em Old} patterns like those shown in columns 1 to 11 in the figure.

\end{itemize}

Because of the intimate connection between information compression and concepts of prediction and probability \cite{li_vitanyi_2009}, the SP system is intrinsically probabilistic. Each SP pattern has an associated frequency of occurrence, and probabilities may be calculated for each multiple alignment and for associated inferences (\cite[Section 4.4]{sp_extended_overview}; \cite[Section 3.7]{wolff_2006}).

The system is realised in the form of a computer model which provides all the examples of multiple alignment in this article. At present, the model works only with one-dimensional patterns \cite[Section 3.3]{sp_extended_overview} but it is envisaged that, at some stage, it will be generalised to work with patterns in two dimensions.

It is intended the SP computer model will be the basis for the development of a high-parallel {\em SP machine}, an expression of the SP theory, a vehicle for research, and a means for the theory to be applied \cite[Section 3.2]{sp_extended_overview}.

Although the main emphasis in the SP programme has been on the development of abstract concepts as outlined above, several of those concepts may be realised in terms of neurons and their inter-connections. This aspect of the SP theory---called {\em SP-neural}---is described in \cite[Section 14]{sp_extended_overview} and \cite[Chapter 11]{wolff_2006}.

The theory has things to say about several aspects of computing and cognition, including unsupervised learning, concepts of computing, aspects of mathematics and logic, the representation of knowledge, natural language processing, pattern recognition, several kinds of reasoning, information storage and retrieval, planning and problem solving, and aspects of neuroscience and of human perception and cognition.

\section{Low-level perceptual features}\label{low_level_features_section_section}

It is now widely accepted that, at `low' levels in vertebrate and invertebrate visual systems, there are processes that recognise perceptual features such as edges and corners. Some relevant evidence is outlined in subsections below.

In this section, the main focus is on features that may be regarded as `explicit' because they derive directly from visual input. But it is well known that we may `see' things that have little or no counterpart in the visual input, such as the `subjective contours' in \cite[Figure 2-6]{marr_2010} or the edge of one leaf where it overlaps another in \cite[Figure 4-1 (a)]{marr_2010}. These kinds of `implicit' features will be considered in Section \ref{things_not_there_section}.

\begin{figure}[!hbt]
\centering
\includegraphics[width=0.5\textwidth]{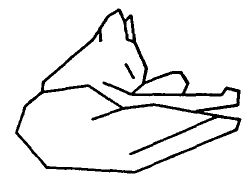}
\caption{Drawing made by abstracting 38 points of maximum curvature from the contours of a sleeping cat, and connecting these points appropriately with a straight edge. Reproduced from Figure 3 in \protect\cite{attneave_1954}, with permission.}
\label{sleeping_cat_figure}
\end{figure}

In two respects, explicit perceptual features sit comfortably with the SP theory:

\begin{itemize}

\item They may be seen to provide a means of encoding perceptual information in an economical manner. For example, Attneave \cite{attneave_1954} writes that ``Common objects may be represented with great economy, and fairly striking fidelity, by copying the points at which their contours change direction maximally, and then connecting these points appropriately with a straight edge.'' (p. 185). He illustrates this with the now-famous picture of a sleeping cat, reproduced in Figure \ref{sleeping_cat_figure}.

\item At lowish levels, perceptual features may function as if they were the atomic symbols that provide the foundation for all higher-level structures, even though they themselves have been constructed from lower-level components.

\end{itemize}

As just indicated, vision begins with images, not perceptual features. The latter must be somehow discovered or detected within the images. The following subsections consider how the SP theory may be applied in this area, starting with a consideration of options for the encoding of light intensities.

\subsection{The encoding of light intensities}

In the design of artificial systems for vision, it seems natural and obvious that light intensities in images should be expressed as numbers. But, in itself, the SP system recognises only atomic symbols, each one of which can be matched in an all-or-nothing manner with another atomic symbol. It is true that, in principle, it may be supplied with patterns that express Peano's axioms or similar information, and it may then interpret numbers correctly \cite[Chapter 10]{wolff_2006}. But this has not yet been explored in any depth and, in any case, numbers are probably a distraction in understanding how SP principles may be applied to vision.

To simplify the discussion here, we shall assume that we are processing monochrome images with just two categories of pixel: black and white. With that kind of representation, the lightness in any given small area may be encoded via the {\em densities} of black and white pixels in that area, without using explicit numbers, somewhat like the encoding of dark and light in monochrome newspaper photographs, at least as they used to be (see also \cite[Section 2.2.3]{wolff_2006}). It is true that such pixels may be represented with the symbols `1' and `0' but these are simply atomic symbols (as required by the SP system), without numerical meanings.

\subsection{Edge detection with neurons}\label{edge_detection_with_neurons_section}

\begin{figure}[!hbt]
\centering
\includegraphics[width=0.9\textwidth]{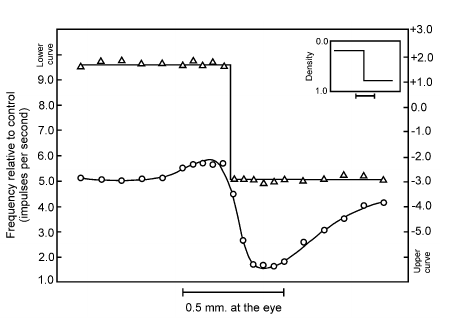}
\caption{Two sets of recordings from a single ommatidium of {\em Limulus} \protect\cite[p. 1248]{ratliff_hartline_1959}. Reproduced from Figure 4, {\em The Journal of General Physiology, 42}, p. 1248, by copyright permission of The Rockefeller University Press.}
\label{limulus_figure}
\end{figure}

It is relevant to this discussion to consider briefly how edges may be detected with neurons. Figure \ref{limulus_figure} shows two sets of recordings from a single visual receptor (`ommatidium') of the horseshoe crab, {\em Limulus polyphemus}. In both sets of recordings, the eye of the crab was illuminated in a rectangular area bordered by a dark rectangle of the same size (producing a step function as shown at the top right of the figure). In both cases, successive recordings were taken with the pair of rectangles in successive positions across the eye along a line which is at right angles to the boundary between light and dark areas. This achieves the same effect as---but is easier to implement than---keeping the two rectangles in one position and taking recordings from a range of receptors across the light and dark areas.

In the top set of recordings (triangles) all the ommatidia except the one from which recordings were being taken were masked from receiving any light. In this case, the target receptor responds with frequent impulses when the light is bright and at a sharply lower rate in the dark. In the bottom set of recordings (circles) the mask was removed so that all the ommatidia were exposed to the pattern of light and dark rectangles. In this case, positive and negative responses are exaggerated near the border between light and dark areas but the target receptor fires at or near a background rate in areas which are evenly illuminated (either light or dark). This kind of effect---which is seen elsewhere in the animal kingdom---appears to be due to lateral inhibition between neurons in the visual system \cite[pp 172-174]{von_bekesy_1967}.

It has been recognised for some time that the dampening of the response in regions of uniform illumination (light or dark) may be seen to achieve the effect of compressing visual information by extracting redundancy from it \cite{barlow_1959}. It is somewhat like the `run-length coding' technique for compression of information: a symbol or group of symbols that repeats in a contiguous sequence may be reduced to a single instance, perhaps marked for repetition.\footnote{See, for example, `Run-length encoding', Wikipedia, \href{http://bit.ly/eyxlY}{bit.ly/eyxlY}, retrieved 2013-02-04.}

A boundary between one uniform area and another may be represented economically by two such compressed representations, side-by-side. In the neural case, the upswing near the light/dark boundary may be seen as an economical representation of the idea that the whole of the preceding area is light, the downswing on the other side may be seen as a succinct marking of the fact that the following area is dark, while the two together may be seen to serve as a compressed representation of the boundary.

Although it is less directly relevant to the present discussion, it is pertinent to mention that there are `complex' cells in mammalian visual systems that respond selectively to edges, and also to `lines' and `slits' (see, for example, \cite[pp 215-219]{frisby_stone_2010}).

\subsection{Edge detection with the SP system}\label{recursive_run-length_coding_section}

In the SP framework, the effect of run-length coding may be achieved via recursion, as illustrated in Figure \ref{recursive_run-length_coding_figure}.\footnote{Compared with the multiple alignment in Figure \ref{winds_from_west_figure}, this one has been rotated by $90\degree$, replacing columns with rows. The two styles are equivalent. The choice between them depends largely on what fits best on the page.}

\begin{figure}[!hbt]
\fontsize{10.00pt}{12.00pt}
\centering
{\bf
\begin{BVerbatim}
0     a b c     a b c     a b c     a b c                  0
      | | |     | | |     | | |     | | |
1     | | | X 1 a b c X   | | |     | | |         #X #X    1
      | | | |         |   | | |     | | |         |  |
2 X 1 a b c X         |   | | |     | | |         |  #X #X 2
                      |   | | |     | | |         |
3                     X 1 a b c X   | | |      #X #X       3
                                |   | | |      |
4                               X 1 a b c X #X #X          4
\end{BVerbatim}
}
\caption{A multiple alignment produced by the SP model with the New pattern `\texttt{a b c a b c a b c a b c}' in row 0 and an appearance of the Old pattern, `\texttt{X 1 a b c X \#X \#X}' in each of rows 1 to 4.}
\label{recursive_run-length_coding_figure}
\end{figure}

Here, each instance of `\texttt{a b c}' in the New pattern in row 0 is matched to an appearance\footnote{In the SP framework, any Old pattern may appear more than once in a multiple alignment. Here, an {\em appearance} of a pattern is {\em not} the same as an {\em instance} of a pattern, as explained in \cite[Section 3.4.6]{wolff_2006}.} of the self-referential Old pattern `\texttt{X 1 a b c X \#X \#X}' in each of rows 1 to 4. It is self-referential because `\texttt{X \#X}' in the body of the pattern may be matched and unified with `\texttt{X~...~\#X}' at the start and end of the pattern.

The encoding of the New pattern that we may derive from this multiple alignment is the relatively short sequence `\texttt{X 1 1 1 1 \#X}'.\footnote{For a description of the method of deriving an encoding from a multiple alignment, see \cite[Section 4.1]{sp_extended_overview} or \cite[Section 3.5]{wolff_2006}.} It achieves lossless compression of the original sequence by recording that the run contains 4 instances of the pattern `\texttt{a b c}', in the manner of unary arithmetic. With lossy compression, the encoding may be reduced to `\texttt{X \#X}', which simply records a sequence of instances of `\texttt{a b c}' without specifying the length of the sequence.

As before, two such encodings, side-by-side, would be an economical representation of the boundary between one uniform region and another.

Of course, this does not look much like lateral inhibition with neurons, as outlined in Section \ref{edge_detection_with_neurons_section}. But at an abstract level, the two things may be seen to produce the same result: the extraction of redundancy from uniform regions, leaving information about the boundaries between such regions as an economical representation of the raw data, like David Marr's `primal sketch' \cite{marr_2010}.

With other developments---such as the generalisation of the SP concepts to two dimensions (Section \ref{sp_outline_section})---this kind of technique may be applied in computer vision.

\subsection{Orientations, lengths, and corners}\label{orientations_lengths_corners_section}

So far, we have said nothing about the orientations of edges or their lengths. In principle, those things may be encoded mathematically, and very economically, in the manner of vector graphics. But that does not seem very likely in a biological system and it is not necessarily the best option for any artificial system that aspires to human-like capabilities in vision (Section \ref{unifying_raster_vector_graphics_section}).

As mentioned above, the visual cortex in mammals is populated by large numbers of `complex' neurons, each one of which responds to an `edge', `slit', or `line', at a particular orientation. There is a good coverage of different angles within each small area (see, for example, \cite[Chapter 9]{frisby_stone_2010}). These observations suggests that, in natural vision, the orientation of any edge may be encoded quite simply and directly in terms of the corresponding type of neuron, and likewise in an artificial system.

A sequence of such codes would describe both the orientation and length of a straight line but it would contain the same kind of redundancy as is discussed in Section \ref{recursive_run-length_coding_section} because the orientation is repeated in successive parts of the line. So we may guess that, in natural vision, some kind of run-length coding may operate, reducing the redundancy within the body of the line and preserving information where the repetition stops---at the points where the line begins and where it ends (see also \cite[Section 13.2.1.4]{wolff_2006}).

This kind of technique may be applied not only to straight lines but also to lines with a uniform curviture. Any such line may be encoded as repeated instances of a short segment that expresses the curvature of the whole line.

With regard to straight lines, some relevant evidence comes from studies showing the existence of `end stopped' hypercomplex cells that respond selectively to a bar of a defined length, or a corner (see, for example, \cite[pp 216-217]{frisby_stone_2010}). In keeping with Attneave's remarks quoted earlier \cite{attneave_1954}, we may guess that, in mammalian vision, the orientation and length of an edge, slit or line, is to a large extent encoded via neurons that record the beginning and end of the line and any associated corners. Orientation-sensitive neurons would provide the input for this `higher' level of encoding.

In artificial systems, this kind of coding may in principle be done within the multiple alignment framework, as outlined in Section \ref{recursive_run-length_coding_section}.

\subsection{Noisy data and low-level features}\label{noisy_data_low-level_features_section}

Readers may, with some justice, object that real visual data is rarely as clean as the example in Figure \ref{recursive_run-length_coding_figure} may suggest. With monochrome images, it is likely that most areas will be some shade of grey, not purely black or purely white, and there are likely to be blots and smudges of various kinds.

What appears to be a promising answer to this kind of problem is that the SP system is designed to search for optimal solutions and is not unduly disturbed by errors of omission, commission and substitution. There is more on this topic in Sections \ref{noisy_data_parsing_recognition_section} and \ref{noisy_data_learning_section}.

\section{Object recognition and scene analysis}\label{object_scene_section}

In some respects, object recognition is like parsing in natural language processing (see, for example, \cite{farabet_etal_2013,han_2005}). Since the SP system works well in parsing, as outlined in \cite[Section 4]{sp_extended_overview}, it may also prove useful in computer vision. Naturally, it would be necessary for the SP machine to have been generalised to work with patterns in two dimensions (Section \ref{sp_outline_section}). And in this discussion we shall assume that low-level perceptual features have been identified, and that they may be treated as atomic symbols, in accordance with the SP theory (Section \ref{low_level_features_section_section}).

Figure \ref{object_recognition_figure} shows schematically how someone's face, with their ears, may be parsed within the multiple alignment framework. Row 0 in the figure contains a New pattern representing incoming information. Each part has been aligned with an Old pattern representing stored knowledge of the structure of an ear, an eye, etc. And these are aligned with a pattern in row 2 representing the higher-level structure of someone's head.

\begin{figure}[!hbt]
\fontsize{09.00pt}{10.80pt}
\centering
{\bf
\begin{BVerbatim}
0         e a r        e y e        n o s e        e y e        e a r       0
          | | |        | | |        | | | |        | | |        | | |
1     E 1 e a r #E     | | |        | | | |        | | |        | | |       1
      |         |      | | |        | | | |        | | |        | | |
2 H 4 E         #E Y   | | | #Y N   | | | | #N Y   | | | #Y E   | | | #E #H 2
                   |   | | | |  |   | | | | |  |   | | | |  |   | | | |
3                  |   | | | |  N 3 n o s e #N |   | | | |  |   | | | |     3
                   |   | | | |                 |   | | | |  |   | | | |
4                  |   | | | |                 Y 2 e y e #Y |   | | | |     4
                   |   | | | |                              |   | | | |
5                  Y 2 e y e #Y                             |   | | | |     5
                                                            |   | | | |
6                                                           E 1 e a r #E    6
\end{BVerbatim}
}
\caption{A multiple alignment showing schematically how a person's face, with their ears, may be recognised.}
\label{object_recognition_figure}
\end{figure}

Although this is schematic, I believe the approach has potential, as described in the following subsections.

\subsection{Noisy data in parsing and recognition}\label{noisy_data_parsing_recognition_section}

Contrary to the impression one might gain from Figure \ref{object_recognition_figure}, the SP system is quite robust in the face of errors in such tasks as parsing natural language or pattern recognition. This is illustrated in the multiple alignment in Figure \ref{noisy_winds_from_west_figure} where the New pattern in column 0 is the same sentence as in Figure \ref{winds_from_west_figure} but with the omission of the `\texttt{n}' in `\texttt{w i n d s}', the addition of `\texttt{x q}' within the word `\texttt{w e s t}', and the substitution of `\texttt{h}' for `\texttt{t}' in `\texttt{s t r o n g}'. Despite these errors, the best multiple alignment created by the SP model is, as shown in the figure, the one that we judge intuitively to be `correct'.

\begin{figure}[!htbp]
\fontsize{06.00pt}{07.20pt}
\centering
{\bf
\begin{BVerbatim}
0   1    2     3    4     5    6     7    8    9     10   11

                                               S
                                               Num ------ Num
                                                          PL
                                               ; -------- ;
                    NP ----------------------- NP
                    0
               N -- N
               Np --------------------------------------- Np
               7
w ------------ w
i ------------ i
               n
d ------------ d
s ------------ s
               #N - #N
         Q -------- Q
         1
    P -- P
    15
f - f
r - r
o - o
m - m
    #P - #P
         NP ------------------ NP
                               0a
                               D --- D
                                     17
t ---------------------------------- t
h ---------------------------------- h
e ---------------------------------- e
                               #D -- #D
                          N -- N
                          Ns
                          4
w ----------------------- w
e ----------------------- e
s ----------------------- s
x
q
t ----------------------- t
                          #N - #N
         #NP ----------------- #NP
         #Q ------- #Q
                    #NP ---------------------- #NP
                                               V --- V
                                                     Vp - Vp
                                                     11
a -------------------------------------------------- a
r -------------------------------------------------- r
e -------------------------------------------------- e
                                               #V -- #V
                                          A -- A
                                          19
s --------------------------------------- s
h                                         t
r --------------------------------------- r
o --------------------------------------- o
n --------------------------------------- n
g --------------------------------------- g
                                          #A - #A
                                               #S

0   1    2     3    4     5    6     7    8    9     10   11
\end{BVerbatim}
}
\caption{A multiple alignment created like the one shown in Figure \ref{winds_from_west_figure} but with errors of omission, commission and substitution in the New pattern in column 0 (representing a sentence to be parsed).}
\label{noisy_winds_from_west_figure}
\end{figure}

This kind of ability to cope gracefully with noisy data is really essential in any system which aspires to explain or emulate our ability to recognise things despite fog, snow, falling leaves, or other things that may obstruct our view.

\subsection{Family-resemblance or polythetic concepts}

A related idea is that the SP system can accommodate `family-resemblance' or {\em polythetic} concepts, meaning that recognition does not depend on the presence or absence of any particular feature or combination of features \cite[Section 9]{sp_extended_overview}; \cite[Section 6.4.3]{wolff_2006}. This is partly because the system can accommodate errors via its search for optimal solutions (Sections \ref{noisy_data_parsing_recognition_section} and \ref{noisy_data_learning_section}), and partly because it allows for the specification of knowledge structures that may have alternatives at any or all points in any given structure.

Most of our concepts are polythetic. Although the possession of four legs seems to be a defining feature of the concept `dog', we would still recognise Fido as a dog, even if he had had the misfortune to lose one of his legs. Likewise for most attributes of most concepts.

As with noisy data, the ability to accommodate polythetic concepts is really essential in any system that aspires to human-like vision.

\subsection{Part-whole hierarchies, class hierarchies, and their integration}\label{part-whole_class-inclusion_section}

A strength of the multiple alignment concept is that it provides a simple but effective vehicle for the representation and processing of part-whole hierarchies, class hierarchies, and their integration, as described and illustrated in \cite[Section 9.1]{sp_extended_overview} and \cite[Section 6.4.1]{wolff_2006}.

The example in \cite[Figure 16]{sp_extended_overview}---a multiple alignment that integrates botanical categories with the parts and sub-parts of a plant---does not describe the visual appearance of an object, but it should be apparent that this system, when it has been generalised to work with patterns in two dimensions, has potential as a means of representing and processing both the parts and sub-parts of an object's image, and how that information relates to any hierarchy of classes to which that object belongs. Each of those two types of hierarchy is an effective means of expressing visual information in a compressed form.

\subsection{Scene analysis}\label{scene_analysis_section}

Scene analysis may also be viewed as a kind of parsing (see, for example, \cite{she_1983}). For the analysis of a seascape, for example, there may be a high-level structure recording the kinds of things that one sees in a typical seascape (sea, beach, sky, rocks, boats, and so on), with a more detailed description for each one of those things.

There seem to be two main complications in scene analysis:

\begin{itemize}

\item Any one thing may be partially obscured by another. In our seascape, a boat may be partially obscured by, for example, waves, sea birds, or members of the crew.

\item The locations of things may be quite variable. A boat may be in the sea or on the beach; people can appear almost anywhere; and so on.

\end{itemize}

Of course, people cope easily with both those things, but there may be a problem with `naive' kinds of parsing system. The SP framework may accommodate these aspects of scene analysis in three main ways:

\begin{itemize}

\item As we saw in Section \ref{noisy_data_parsing_recognition_section}, parsing can be done successfully despite errors or omission, commission, or substitution. Thus there is reason to believe that, when the SP models have been generalised to work with patterns in two dimensions, an object may be recognised even if it is partially obscured.

\item The variability of scenes is broadly similar to the variability of sentences in natural language. Artificial parsing systems, including the SP system, can cope with that variability by providing information about a wide variety of types of sentences and phrases, including recursive forms such as {\em This is the man all tattered and torn that kissed the maiden all forlorn that milked the cow with the crumpled horn ...}. The same principles may be applied to vision.

\item Where existing knowledge can't cope, the system may learn---as discussed in Section \ref{unsupervised_learning_section}, next.

\end{itemize}

\section{Unsupervised learning and the discovery of objects and classes}\label{unsupervised_learning_section}

There is clearly a close relationship between learning and vision since vision is an important means of gaining new information about the world. In general, we learn via vision in a manner that is `unsupervised' in the sense that it does not require the intervention of a `teacher', or the provision of `negative' samples, or the grading of samples from simple to complex ({\em cf.}~\cite{gold_1967}). We take in information through our eyes (and other senses) and try to make sense of it as best we can.

In this section, we consider unsupervised learning as it has been developed in the SP framework, and how it may be applied in vision. As a preliminary to this discussion, readers may like to consult \cite[Section 5]{sp_extended_overview}, which outlines how unsupervised learning is done in the SP model. In particular, the `DONSVIC' principle---how `natural' structures may be discovered via information compression---is described in \cite[Section 5.2]{sp_extended_overview}.

In brief, the product of learning from a body of information, {\bf I}, may be seen to comprise: a {\em grammar} ({\bf G}) which captures {\em redundancies} in {\bf I}---information that is repeated within {\bf I}---and may be seen as a distillation of the `essence' of {\bf I}; and an {\em encoding} ({\bf E}) which expresses the non-redundant features of {\bf I}. In accordance with the principle of {\em minimum length encoding} \cite{solomonoff_1964}, the SP system aims to minimise the overall size of {\bf G} and {\bf E}.

\subsection{The discovery of objects via stereo matching}\label{discovery_via_stereo_matching_section}

\begin{figure}[!hbt]
\centering
\includegraphics[width=0.9\textwidth]{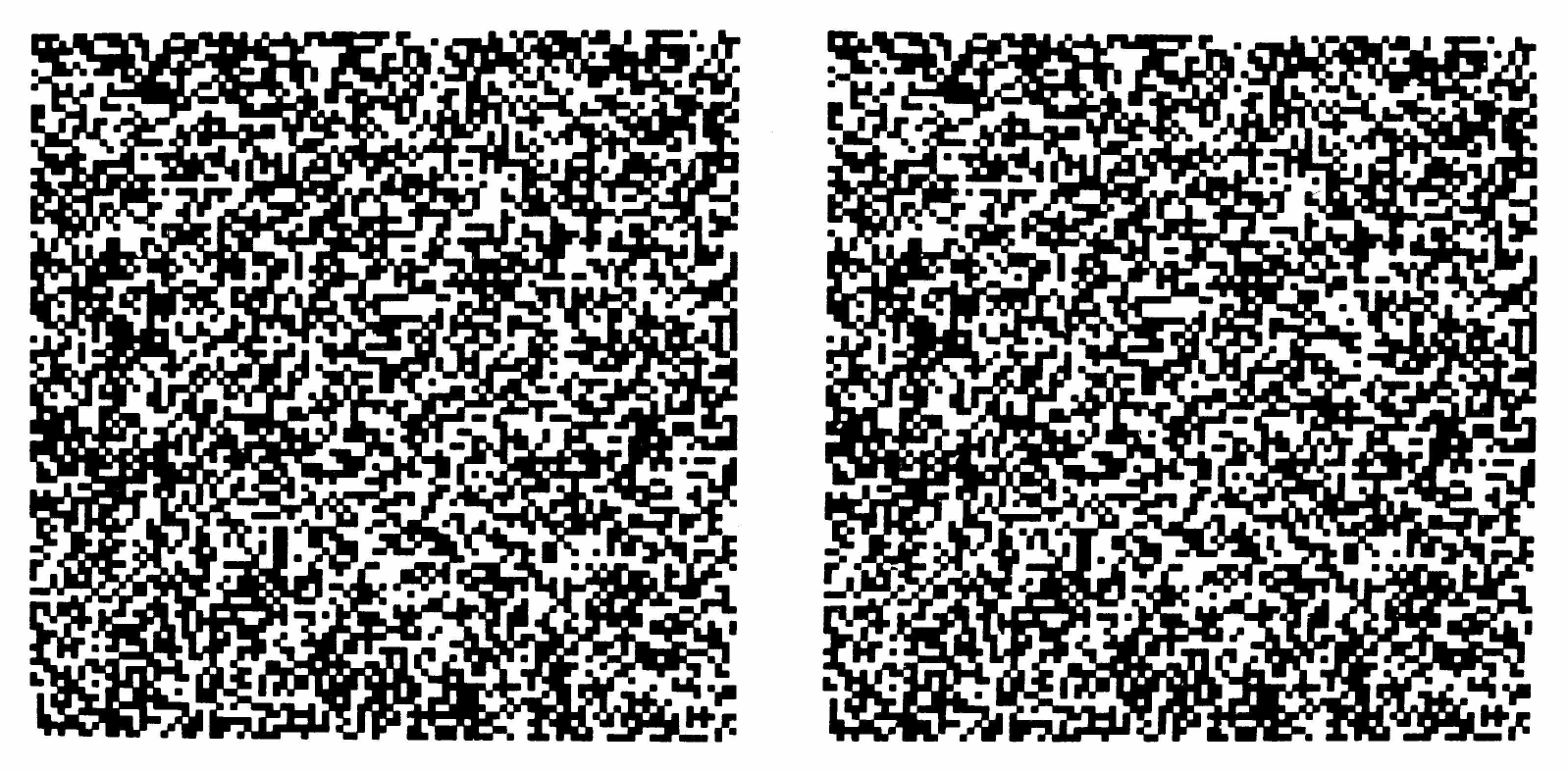}
\caption{A random-dot stereogram from \protect\cite[Figure 2.4-1]{julesz_1971}, reproduced with permission of Alcatel-Lucent/Bell Labs.}
\label{stereogram_1_figure}
\end{figure}

As with the structures of natural language, it is clear that we have to learn the structures that are significant in vision, including objects.\footnote{The Chomskian doctrine that children `acquire' their native language or languages via an inborn knowledge of `universal grammar' depends on the still-unproven idea that such a grammar can be defined for all the world's languages, and it is still not clear how the acquisition process might work.} Some insights into how this may be done may be gained from a consideration of random-dot stereograms like the one shown in Figure \ref{stereogram_1_figure}.

\begin{figure}[!hbt]
\centering
\includegraphics[width=0.9\textwidth]{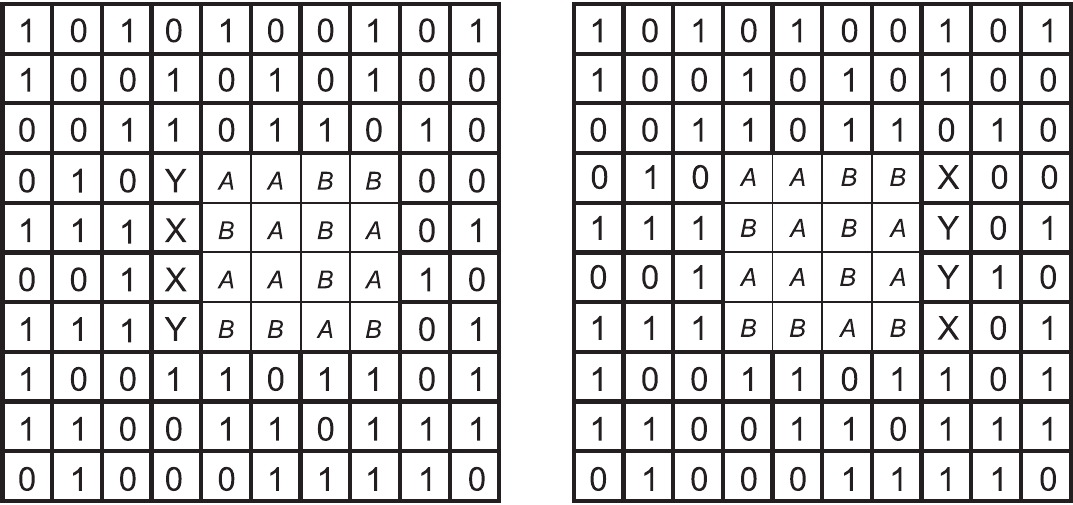}
\caption{Diagram to show the relationship between the left and right images in Figure \ref{stereogram_1_figure}. Reproduced from \protect\cite[Figure 2.4-3]{julesz_1971}, with permission of Alcatel-Lucent/Bell Labs.}
\label{stereogram_2_figure}
\end{figure}

Here, each of the two images is a random array of black and white pixels, with no discernable structure. But there is a relationship between them, as shown in Figure \ref{stereogram_2_figure}: both images are the same except that a square area near the middle of the left image is further to the left in the right image.

When these images are viewed in a stereoscope (so that the left image is viewed by the left eye and the right image by the right eye), the central square appears as a discrete object suspended above the background.\footnote{Some people are able to see the square by viewing the images directly, with some defocussing to help merge them into one.} The focus of interest here will be on how we come to see that discrete object, while possible implications for our understanding of depth perception are discussed in Section \ref{depth_stereoscopic_vision_section}.

A little analysis shows that seeing the central square means finding an alignment between pixels in the left image and pixels in the right image, that there are many alternative such alignments, and that some are better than others. One solution is the algorithm developed by Marr and Poggio \cite{marr_poggio_1979}. Another potential solution is the kind of processing that builds multiple alignments in the SP models, but generalised for two dimensions. As noted in \cite[Section 4.3]{sp_extended_overview}, the complexity of the matching problem can, in general, be reduced by applying constraints to the process of searching and thus reducing the size of the search space.

Figure \ref{julesz_analogue_figure} shows how the SP model can solve a one-dimensional analogue of the stereo matching problem. Here, the Old pattern (row 1) may be seen as an analogue of the left image and the New pattern (row 0) may be seen to stand in for the right image. Both patterns have been prepared from a random sequence of digits,\footnote{The random sequence of digits, with values between 0 and 9, inclusive, was generated by the Random Integer Generator from Random.org (\href{http://www.random.org}{www.random.org}). The results are, they say, better than with pseudo-random number algorithms because ``atmospheric noise'' is the source of randomness.} with a displacement of the middle section, much as in Figure \ref{stereogram_2_figure}. This multiple alignment is the best of several different multiple alignments created by the SP model with those two patterns.

\begin{figure}[!hbt]
\fontsize{08.50pt}{10.20pt}
\centering
{\bf
\begin{BVerbatim}
0     4 7 4 6 4 1 3 7 5     8 5 2 4 0 2 9 1 9 3 8 0 1 4 1 1 2 9 7 1 2    0
      | | | | | | | | |     | | | | | | | | | |     | | | | | | | | |
1 J a 4 7 4 6 4 1 3 7 5 9 4 8 5 2 4 0 2 9 1 9 3     1 4 1 1 2 9 7 1 2 #J 1
\end{BVerbatim}
}
\caption{The best multiple alignment created by the SP computer model with an Old pattern (row 1) and a New pattern (row 0) as one-dimensional analogues of the left and right images in a random-dot stereogram.}
\label{julesz_analogue_figure}
\end{figure}

In the figure, one can see how the central sequence of 10 integers (analogous to the central square in Figure \ref{stereogram_2_figure}) has been isolated from the `background' sequences to the left and right, and this despite repetitions of integers in both patterns and the formation of plenty of `wrong' alignments on the route to the `correct' result. It seems likely that the processes can be generalised to work with patterns in two dimensions.

\subsection{Structure from motion}\label{structure_from_motion_section}

The kinds of processing just described may also be applied to objects in motion.

Consider, for example, a flatfish with a sandy, speckled colouration, lying on a sandy and speckled area on the bed of the sea. Such a creature would be very well camouflaged but with one proviso: it must stay still. As soon as it moves, it will become very much easier to see. Why? Apart from the motion itself, an important reason seems to be that movement creates at least two images (normally more), rather like the two images in a random-dot stereogram. And by a process of matching, much as described above, a predator or other observer will be able to see the fish standing out as a distinct entity with distinct boundaries---like the square that can be seen when the two images in Figure \ref{stereogram_1_figure} are viewed in a stereoscope.

More generally, we see any object in motion---such as a car travelling along a road---as a single entity, not a multitude of images like the frames in a video or film. In all such cases, we merge the many instances into one, and likewise for the background. The process of merging those many instances, which is likely to yield high levels of compression, requires a process of matching and unification, much as before. And those processes serve to define the boundaries of the entity and to distinguish it from the background.\footnote{Of course, there may be additional factors, such as changing perspectives as the moving object passes the observer, but the principles that have been outlined would still apply.}

\subsection{Motion and speed}

\begin{figure}[!hbt]
\centering
\includegraphics[width=0.9\textwidth]{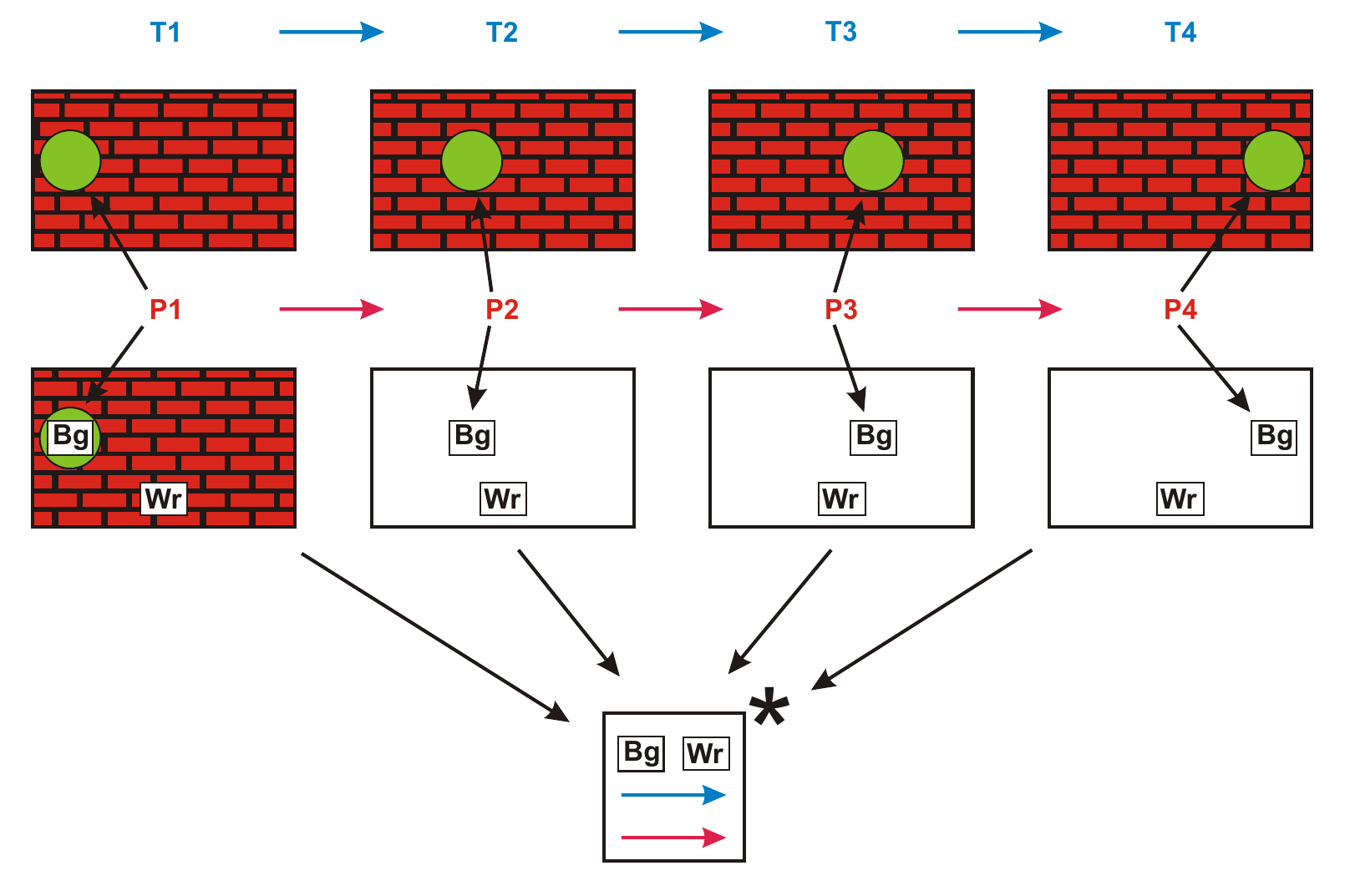}
\caption{A schematic representation of how successive images of a moving ball, with a wall in the background, may be compressed, with an encoding of the ball's motion and its speed. {\em Key}: {\bf T1} to {\bf T4}---times for the corresponding images; {\bf P1} to {\bf P4}---successive positions of the ball; {\bf Blue arrows}---each one represents a time interval between one frame and the next, with a past-to-future direction; {\bf Red arrows}---each one represents the distance travelled by the ball between one frame and the next, with the direction of travel; {\bf Bg}---a relatively short name or code for the ball, with an associated colour (green); {\bf Wr}---a relatively short name or code for the wall, with an associated colour (red); {\bf *}---iteration of the encoding in the envelope.}
\label{moving_object_figure}
\end{figure}

What about the motion itself, and the associated concept of speed? Figure \ref{moving_object_figure} is intended to suggest how these things may be encoded from successive images of a ball moving horizontally in front of a wall.

The raw data is shown in the top four frames with the time for each frame ({\bf T1} to {\bf T4}) marked above them, and with the position of the ball in each frame marked with {\bf P1} to {\bf P4}.

In the first of the lower four frames, the four images of the ball have been merged (unified) into a single image, with `{\bf Bg}' (short for `ball (green)') as its label or code, in accordance with the {\em schema-plus-correction} technique for information compression (\cite[Section 3]{sp_foundations}; \cite[Chapter 2]{wolff_2006}).\footnote{In this discussion, we shall ignore any compression that may be achieved by taking advantage of uniformity and corresponding redundancy within the image of the ball itself, as discussed in Section \ref{low_level_features_section_section}.} In the following 3 frames, the code for the ball (`{\bf Bg}') replaces the image of the ball itself.\footnote{The relatively large size of `{\bf Bg}' in the figure, with its surrounding envelope, may suggest that it does not yield much compression of the image of the ball. But it has been shown quite large simply to ensure that it is readable. Likewise for `{\bf Wr}' and its envelope. In practice, it is likely that labels like these would be encoded with far fewer bits than would be required for the things they represent.}

In a similar way, in that first frame, the 4 images of the background have been unified and marked with `{\bf Wr}' (short for `wall (red)'), and in the following 3 frames, {\bf Wr} serves to represent the wall.\footnote{As with the ball, we shall ignore any compression that may be achieved via the detection and extraction of redundancy within uniform areas within the image of the wall.}

Further compression is possible because there is repetition of the time interval between one frame and the next (marked in the figure with blue arrows), and likewise for the distance that the ball has travelled in each time interval (marked in the figure with red arrows). In each case, the repeated patterns may be merged to create a single instance.

The overall result would be something like the following:

\begin{itemize}

\item The unified images of the ball and the wall, each with their associated codes, `{\bf Bg}' and `{\bf Wr}', respectively.

\item An encoding of the original four frames and their associated information, something like this: `({\bf Bg}, {\bf Wr}, ${\bm{\textcolor{blue}\rightarrow}}$, ${\bm{\textcolor{red}\rightarrow}}$)\textsuperscript{*}', with the superscript `*' to mark iteration.\footnote{Some additional information may be needed to show inter-relationships amongst the symbols}.

\end{itemize}

In accordance with the distinction between `grammar' and `encoding' (Section \ref{unsupervised_learning_section}, above; see also \cite[Section 6.8]{sp_benefits_apps}; \cite[Section IV-A]{sp_big_data}), the first item may be regarded as a grammar for the original data, while the second item may be regarded as an encoding of those data in terms of the grammar. The grammar and the encoding is a compressed representation of the ball, the wall, and the left-to-right motion of the ball at a uniform speed. Further refinement would be needed if the speed of the ball was changing or if the path of the ball was curved.

The waterfall illusion\footnote{See ``Motion aftereffect'', Wikipedia, \href{http://bit.ly/1l6nX5g}{bit.ly/1l6nX5g}, retrieved 2014-03-20.} suggests that we may have more than one system for assessing motion and speed: a quick-and-dirty system for when fast responses are needed and a slower-but-more-accurate system which may serve us better when there is less time pressure.

\subsection{Deriving concepts from fragments}\label{concepts_from_fragments_section}

If we only ever see parts of an object---perhaps a rare creature in its natural habitat that we have only seen in fleeting glimpses---we can nevertheless develop a coherent concept of the whole object via alignments amongst the fragmentary views: `\texttt{A B}' may be aligned with `\texttt{B C}' and unified to create `\texttt{A B C}'; `\texttt{C D}' may be aligned with `\texttt{D E}' to create `\texttt{C D E}'; `\texttt{A B C}' may be aligned with `\texttt{C D E}' ..., and so on. This is like the `sequence assembly' technique in bioinformatics,\footnote{See ``Sequence assembly'', Wikipedia, \href{http://bit.ly/1CsZOvi}{bit.ly/1CsZOvi}, retrieved 2013-02-21.} or the stitching together of overlapping photos to create a panorama.

Similar things may be said about how we develop a coherent view of what's in front of us despite frequent saccadic movements of our eyes from one focus of attention to another. The several views may be stitched together to create a single view.

In both cases, the matching may be achieved via multiple alignment, as developed in the SP theory.

\subsection{The discovery of classes of entity, hierarchies of classes, and part-whole hierarchies}

Similar things may be said about the learning of categories, classes or concepts like `person' or `house', and hierarchies of categories like `species', `genus', `family', `order', and so on (Section \ref{part-whole_class-inclusion_section}).

Any such category is a powerful aid to information compression because it saves the need to repeat information in individual instances. If we know that something is a house, we may assume that it has a roof, walls, windows, and at least one door---and that it is likely to have a kitchen, bathroom, bedrooms, and so on. As in object-oriented design, lower-level classes may `inherit' attributes from higher-level categories.

There is clear potential with the SP system to find the commonalities amongst collections of entities, to create classes and hierarchies of classes like those that have been mentioned, and to learn the relationship between an entity and its parts. As mentioned earlier, the multiple alignment framework can accommodate the seamless integration of class hierarchies with part-whole hierarchies \cite[Section 9.1]{sp_extended_overview}.

\subsection{Noisy data and learning}\label{noisy_data_learning_section}

As with such things as the parsing of natural language and pattern recognition (Section \ref{noisy_data_parsing_recognition_section}), unsupervised learning via information compression can yield `correct' results despite errors of omission, commission or substitution in the data which is the input for learning (\cite[Section X-B]{sp_big_data}; \cite[Section 5.3]{sp_extended_overview}; \cite[Section 12.6.1]{wolff_2006}).

As mentioned above, the product of learning from a body of information, {\bf I}, may be seen to comprise: a {\em grammar} ({\bf G}) and an {\em encoding} ({\bf E}). As a general rule, errors in {\bf I} are likely to be excluded from {\bf G} for two main reasons:

\begin{itemize}

\item Although errors of addition or substitution can, collectively, be quite common, any particular kind of error is likely to be rare. Since {\bf G} captures the redundancies in {\bf I} and excludes non-redundant information, it is likely exclude most of the errors of addition or substitution in {\bf I}.

\item A consequence of aiming to minimise the overall size of {\bf G} and {\bf E} (the principle of minimum length encoding) is that {\bf G} may generalise beyond the data in {\bf I} without over-generalising. This means that errors of omission may be corrected in {\bf G} (via generalisation) but, normally, the system would not introduce new errors via over-generalisation.

\end{itemize}

\section{Space and depth}\label{space_depth_section}

In the SP theory, all kinds of knowledge are represented with patterns in one or two dimensions. Superficially, this seems to rule out anything with more dimensions, and suggests that there might be a need to introduce patterns with three dimensions and possibly more. However, this has been rejected, at least for the time being, for these main reasons:

\begin{itemize}

\item Although the multiple alignment concept may in principle be generalised to patterns in three or more dimensions, it is difficult to see how it could be made to work in practice and it looks implausible as a model for any kind of structure or process in the brain.

\item A central idea in SP-neural (Section \ref{sp_outline_section}) is that the cerebral cortex---which is, topologically, a two-dimensional sheet---may be, in some respects, like a sheet of paper on which {\em pattern assemblies} (neural analogues of SP patterns) may be written \cite[Chapter 11]{wolff_2006}. This is shown schematically in \cite[Figure 31]{sp_extended_overview}.

\item If we exclude processes of interpretation in terms of harmonics, colours, or the like, raw sensory data may be seen to come in either one dimension (eg sound) or two (eg visual images).

\item Three-dimensional structures may be represented with patterns in two dimensions, somewhat in the manner of architects' drawings \cite[Section 13.2.2]{wolff_2006}. With the development of mathematical concepts within the SP framework \cite[Chapter 10]{wolff_2006}, four or more dimensions may be represented in much the same way as is done now with mathematical techniques.

\end{itemize}

The following three subsections consider some aspects of the visual perception of space and depth, and how the SP theory may be applied.

\subsection{Three-dimensional objects}\label{three_dimensional_objects_section}

\begin{figure}[!hbt]
\centering
\includegraphics[width=0.6\textwidth]{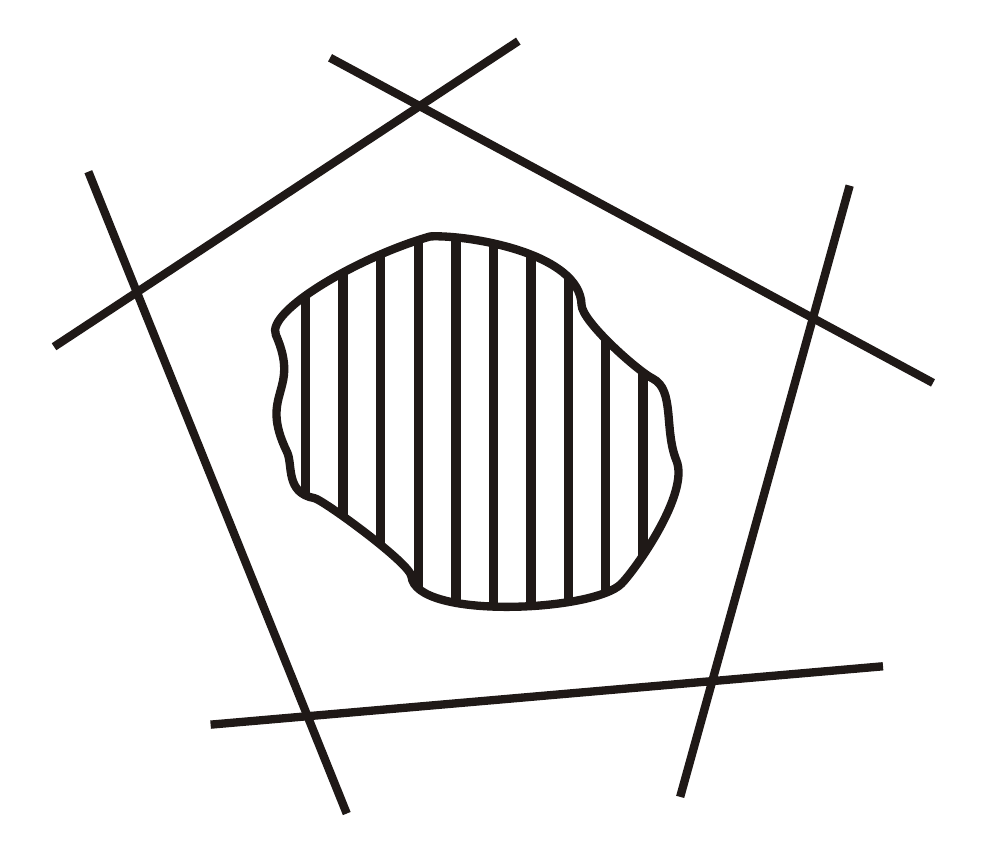}
\caption{Plan view of a 3D object, with each of the five lines around it representing a view of the object, as seen from the side.}
\label{3d_object_views_figure}
\end{figure}

If an object is viewed from several different angles, with overlap between one view and the next (as illustrated in Figure \ref{3d_object_views_figure}), the several views may be stitched together to create what is at least a partial and approximate 3D model of the object, much as in widely-available systems for creating 3D models from sets of photographs.\footnote{\raggedright See, for example, ``Big Object Base'' (\href{http://bit.ly/1gwuIfa}{bit.ly/1gwuIfa}), ``Camera 3D'' (\href{http://bit.ly/1iSEqZu}{bit.ly/1iSEqZu}), or ``PhotoModeler'' (\href{http://bit.ly/MDj70X}{bit.ly/MDj70X}.)}

This is similar to the piecing together of fragments to create a coherent concept, as outlined in Section \ref{concepts_from_fragments_section}. As before, it may be achieved via multiple alignment as that concept has been developed in the SP theory.

The model will be partial if, for example, it excludes views from above or below. And it is likely to be approximate because a given set of views may not be sufficient for an unambiguous definition of the object's geometry: there may be variations in the shape that would be compatible with the given set of views.

Do these deficiencies matter? For many practical purposes, the answer is likely to be ``no''. If we want a rock to put in a rockery, or a stick to throw for a dog, the exact shape is not important. And if we want more accurate information, we can inspect the object more closely, or supplement vision with touch.

Evidence that people do something like what has been described is our ordinary experience that things can be harder to recognised from unfamiliar viewpoints than from familiar ones---the basis of some trick photos. That observation is confirmed in experimental studies showing that people are both slower at recognising things, and less accurate, when the viewpoint is unfamiliar \cite{tarr_1995,bulthoff_edelman_1992,tarr_pinker_1989}.

Evidence that our vision can be a rather poor guide to 3D structure is the failure of most people to see the distorted nature of the Ames' distorted room (see also Section \ref{triangulation_section}).\footnote{For readers who are not familiar with this illusion, a person looks into one end of a room that appears to have a conventional rectangular form but is actually constructed so that one of the two corners opposite the viewer is stretched away and is relatively high, while the other corner is nearer to the viewer and is relatively low. Anyone standing in the near corner appears to be large, and they appear to shrink if they walk to the far corner. A description of the Ames' room and the illusion may be found in ``Ames room'', Wikipedia, \href{http://bit.ly/1Ct0mkV}{bit.ly/1Ct0mkV},retrieved 2013-03-30.} Somehow, this deficiency in our vision does not prevent us from dealing very effectively with cups, plates, tables, and other everyday objects, or from from finding our way around. It seems likely that, in a similar way, robots may be effective in many situations without a geometrically-accurate knowledge of objects and surroundings.

Although what has been described is like the stitching together of overlapping photos to create a panorama, the SP theory suggests that, with people, the visual information would be compressed via the encoding, within the SP system, of part-whole relations, class-inclusion relations, and other kinds of regularities.\footnote{It is true that, in digital systems, photos are normally compressed via JPEG or similar technique. But, as indicated in \cite[Section 5.2]{sp_extended_overview}, there is potential for the SP system to yield higher levels of compression and more natural structures.}

\subsection{Building a model of one's environment and finding one's way around}

Similar processes may be at work when we move around in our environment and learn about it. Successive views that overlap each other may be stitched together, as before, to create a model of the streets or other places where we have been. This is essentially what is done with Google's ``Street View''.\footnote{See \href{http://bit.ly/1rprOMJ}{bit.ly/1rprOMJ}.} and appears to be the basis for the creation of 3D maps using smartphones in the Google-funded ``Project Tango''.\footnote{See ``Software would make 3-D maps using smartphones'', The Columbus Dispatch, 2014-03-24, \href{http://bit.ly/1gqjZBe}{bit.ly/1gqjZBe}.}

The main difference between what has been achieved with Street View and what is envisaged for the SP system is that, in the latter case, as noted in Section \ref{three_dimensional_objects_section}, visual information would be compressed via the mechanisms in the SP system.

As with objects (Section \ref{three_dimensional_objects_section}), a model of our environment that is created via overlapping views may not be geometrically precise.\footnote{From my own experience of exploring caves, I know that, while one can build up a good knowledge of how different passages connect with each other, one's understanding of their 3D geometry can be hazy, and can lead to some surprises if one has the opportunity to see a 3D model that is based on a proper survey, with measurements of distances and angles.} But, as before, some ambiguity may not matter very much for many practical purposes. Topological maps, such as the classic map of the London underground, can be quite good enough for finding one's way around. However, if greater geometric accuracy is needed, it may be increased by gathering more information, especially information about areas between roads, paths or other routes.

In connection with finding one's way around, the SP system may be relevant in two ways:

\begin{itemize}

\item If a robot has stored representations of one or more places, perhaps compressed via recurrent patterns as indicated in \cite[Section 2.1]{sp_extended_overview}, then, via the building of multiple alignments (as in Section \ref{object_scene_section}), it should be able to recognise when it has reached one of those places, using incoming visual information as New patterns and stored knowledge as Old patterns. If it has stored information about an entire route or network of routes, then, within that environment, it should be able to identify where it is at any time. Similar things may be true of people.

\item With an appropriate set of Old patterns, each one of which represents a direct connection between two places, the SP system, via the building of multiple alignments, can work out one or more routes between any two of the relevant places, including routes via two or more of the direct connections \cite[Chapter 8]{wolff_2006}. The example in Figure \ref{planning_figure} shows one such flying route between Beijing and New York.

\end{itemize}

\begin{figure}[!hbt]
\fontsize{12.00pt}{14.40pt}
\centering
\begin{BVerbatim}
0          1          2           3           4

Beijing ------------------------- Beijing
                                  1
                                  Melbourne - Melbourne
                                              10a
                      Cape_Town ------------- Cape_Town
                      14
           Paris ---- Paris
           25a
New_York - New_York

0          1          2           3           4
\end{BVerbatim}
\caption{A multiple alignment showing a flying route between Beijing and New York, one of several produced by the SP computer model with a set of Old patterns, one for each leg of this and other possible journeys. Reproduced from Figure 8.5 (e) in \cite{wolff_2006}, with permission.}
\label{planning_figure}
\end{figure}

These points about how we may build a model of our environment and find our way around relate to the topic in robotics of `simultaneous localization and mapping' (SLAM).\footnote{See, for example, ``Simultaneous localization and\ mapping'', Wikipedia, \href{http://bit.ly/1ikQTRR}{bit.ly/1ikQTRR}, retrieved 2014-04-04.}

\subsection{Depth perception and stereoscopic vision}\label{depth_stereoscopic_vision_section}

Without attempting a comprehensive discussion of the complex subject of depth perception, this section offers some thoughts about stereoscopic vision, and the possible relevance of the SP theory.

\subsubsection{Triangulation}\label{triangulation_section}

For any given object that we are looking at, we can in principle work out its distance by a process of triangulation like that which has been widely used in cartography, at least as it used to be. But there appear to be snags:

\begin{itemize}

\item For this mechanism to work with reasonable accuracy, it would be necessary for one to have a rather accurate sense of the direction of gaze for each eye and the angle between that direction of gaze and the line between the two eyes. It seems unlikely that we can sense the positions of our eyes with the necessary accuracy.

\item There is evidence that, with the previously-mentioned Ames' distorted room illusion, the illusion persists when people view the room with two eyes \cite{glennerster_etal_2003}, although, in that case, the effect may be reduced \cite{gehringer_engel_1986}. This suggests that any information about distance that may be gained via triangulation, or any other clue such as the focussing of our eyes, is not sufficiently clear or precise to overcome viewers' preconceptions that the room has the conventional rectangular form.

\item Triangulation cannot work with a stereoscope or a 3D film because what we are looking at is all at one distance, with nothing to differentiate one part of the picture from another. The spear which makes us jump as we see it coming out of the cinema screen towards us is no closer to us than anything else in the film.

\end{itemize}

We cannot rule out triangulation altogether---it may have a role in some situations---but some other mechanism is needed to explain how we see depth with a stereoscope or a 3D film.

\subsubsection{Possible alternatives}\label{alternatives_section}

With random-dot stereograms, it is clear that our brains are capable of forming an alignment between the left and right images that is good enough to identify the displaced area in the middle as a discrete entity (Section \ref{discovery_via_stereo_matching_section}). By identifying the displaced area and distinguishing it from the surrounding area, we may also gain an accurate knowledge of the size of the displacement.

What can the size of the displacement (D) tell us about depth? With a knowledge of the distance between the eyes but without other information, it can tell us the {\em ratio} of the distances between the observer and the object (A), and between the object and the background (B). With some other information about the first of these distances, we can infer the second distance, and {\em vice versa}.

But the foregoing presupposes a knowledge of geometry. Since geometry is a relatively recent invention in the millions of years that there have been people on earth, and since a knowledge of geometry would normally require schooling and would in any case not be available to animals, it seems likely that some other mechanism is needed to explain how, in most cases, people and other animals make judgements about depth using observations about displacements.

The tentative answer suggested here is that:

\begin{itemize}

\item From experience, we build a table of associations between the distances A, B, and D.

\item In later situations, a knowledge of any two of those three values would enable us look up the third.

\end{itemize}

Both of these processes are well within the scope of the SP system. The system is designed to discover recurrent patterns in data, including the kinds of association considered here \cite[Section 5]{sp_extended_overview}, \cite[Chapter 9]{wolff_2006}. And table look-up can be modelled quite simply via multiple alignment \cite[Section 3.1]{wolff_sp_intelligent_database}, \cite[Section 6.3]{wolff_2006}.

With those kinds of capability, we have a means of making inferences about depth that may not always be geometrically accurate but may be good enough for many practical purposes.

\section{Some other aspects of vision}

The SP theory has things to say about some other aspects of vision, as discussed in the following subsections.

\subsection{Seeing things that are not there}\label{things_not_there_section}

As noted in Section \ref{low_level_features_section_section}, we often `see' things that are not objectively present in what we are looking at. We may see `subjective contours' in certain kinds of images, or we may see the edge of a leaf where it overlaps another leaf despite there being little or nothing to mark the boundary.

The multiple alignment in Figure \ref{winds_from_west_figure} provides an example of how the SP system may accommodate these kinds of things. Here, the New pattern is the sentence `\texttt{w i n d s f r o m t h e w e s t a r e s t r o n g}' with nothing to mark the boundary between one word and the next. But those boundaries are clearly marked in the figure via the parsing of the sentence into its constituent parts.

More generally, we infer things that are not immediately visible: when we see the unbroken shell of a nut, we expect to find an edible kernel inside; when we see a horse partially obscured by a tree, we expect to see the whole animal when it moves into full view; and so on.

This kind of inference is an integral part of how the SP system works. In Figure \ref{noisy_winds_from_west_figure}, the word `\texttt{w i n d s}' appears in the New pattern as `\texttt{w i d s}', but the parsing interpolates the missing `\texttt{n}'. In an example of pattern recognition \cite[Section 9, Figure 16]{sp_extended_overview}, the SP system identifies an unknown plant as, probably, an example of the Meadow Buttercup ({\em Ranunculus acris}) from a few features in the New information with which it is supplied: that the stem is hairy, that the petals are yellow, that there are numerous stamens, and more. From its stored knowledge about that species, the system can infer several other things, not in the New information: that the plant photosynthesises, that it has five petals, that it is poisonous, and more.

\subsection{Variations in image size}

Variations in the sizes of images are significant in both natural vision (Section \ref{image_size_natural_vision_section}) and in the processing of images by computers (Section \ref{unifying_raster_vector_graphics_section}).

\subsubsection{Recognition despite variations in image size}\label{image_size_natural_vision_section}

A prominent feature of natural vision is that we can recognise something despite wide variations in viewing distance and corresponding variations in the size of the retinal image.\footnote{This is related to but different from the phenomenon of `size constancy': that, within limits, our perception of an object's size remains the same, regardless of viewing distance or the size of the retinal image.} Although this phenomenon is not consistent with any simple pattern-matching model of vision, it appears that it can be accommodated within the SP framework.

Let us suppose that, as described in Section \ref{recursive_run-length_coding_section}, the image to be processed is reduced to a `primal sketch', showing boundaries between uniform areas but without the redundancy within those areas. And let us suppose that the redundancy in any line with a uniform direction or curve has been reduced or eliminated as described in Section \ref{orientations_lengths_corners_section}.

For any given image, the effect of those kinds of processing will be to reduce or eliminate variations in the size of the original image. The encoding that is derived from a large version of the scene will be much the same as the encoding that is derived from a small version.

This kind of capability in the SP system, together with the system's flexibility in matching patterns \cite[Section 4.2]{sp_extended_overview}, should mean human-like capabilities in recognising things despite wide variations in the sizes of images of any given entity.

\subsubsection{Unifying raster graphics and vector graphics}\label{unifying_raster_vector_graphics_section}

With image processing, it is generally understood that raster graphics are good for photographs or paintings but go fuzzy if images are enlarged too much, while vector graphics have the advantage that everything stays sharp when images are enlarged but are really only suitable for things like text and diagrams.

The SP system may provide a means of overcoming this image-processing apartheid. Images of any kind may be encoded economically as described in this article. But unlike raster graphics, images encoded via the SP system should be expandable without loss of definition, much as in vector graphics. The key to this scalability of SP-encoded images appears to be recursive encoding of regions of uniformity---in the manner of run-length coding---applied to areas (Section \ref{recursive_run-length_coding_section}) and also to lines (Section \ref{orientations_lengths_corners_section}). The recursive encoding of areas and lines should allow the whole image to be expanded indefinitely without loss of definition.

Any such encoding would be a useful step forward in the development of a UFK \cite[Section III]{sp_big_data}.

\subsection{Lightness constancy and colour constancy}

Another prominent feature of natural vision is `lightness constancy': the fact that, normally, we perceive the lightness of an object to be fixed, despite wide variations in the intensity of the incident light and corresponding variations in the amount of light that is reflected from the object (its `luminence'). We would normally see a lump of coal as black and snow as white, even though the coal in bright sunlight may be reflecting more light per unit area than snow in shadow.

In order to account for this phenomenon, it seems necessary to suppose that, for each kind of object, we maintain some kind of table of associations between levels of illumination and corresponding values for luminance. Since we are unlikely to have an inborn knowledge of coal, snow, and the like, we must suppose that those tables are learned. As noted in Section \ref{alternatives_section}, learning associations of that kind is part of what the SP system is designed to achieve.

Notice that any given table can only be applied if we have some idea of what kind of object we are looking at, otherwise we might see coal as if it was snow, or {\em vice versa}. There is some evidence that our perception of the lightness of an object does indeed depend on what we think the object is \cite[Chapter 16]{frisby_stone_2010}. In a similar way, our judgements of lightness seem to depend on our perceptions of how a given object is illuminated \cite[Figure 1.10]{stone_2012}.

It seems likely that much of what has been said in this section about lightness constancy would also apply to colour constancy: the way we see the colour of an object to be fixed, despite wide variations in the colour of the incident light and corresponding variations in the colour of the light that is reflected from the object.

In terms of information compression as a unifying principle in computing and cognition, it is pertinent to mention that lightness constancy and colour constancy may each be seen as a means of encoding information economically. It is simpler to remember that a particular object is `black' or `red' than all the complexity of how its appearance changes in different lighting conditions.

\subsection{The role of context in recognition}\label{context_recognition_section}

It is often remarked that we recognise things more easily in their familiar contexts than in unfamiliar ones. This is confirmed in formal studies (see, for example, \cite{bar_ullman_1993,oliva_torralba_2007}).

This observation makes sense in terms of the SP framework because any part of a multiple alignment may be a context for any other, and because of the way the system searches for a global optimum which embraces any given entity and its context. If, in our seascape example (Section \ref{scene_analysis_section}), we see a beach and the sea then, in effect, we are primed to see boats---because, in that context, boats are likely to yield multiple alignments with better scores than, say, office furniture.

Context also has a role in resolving ambiguities, as outlined at the end of the next subsection.

\subsection{Ambiguity in perception}\label{ambiguity_section}

A prominent feature of perception is that, with some kinds of sensory input, there may be more than one plausible interpretation. Examples include the `young woman / old woman' picture and the `duck / rabbit' picture of psychology text books. An example with natural language is the ambiguous sentence {\em Fruit flies like a banana}, the second part of {\em Time flies like an arrow. Fruit flies like a banana}, attributed to Groucho Marx.

In the SP framework, this kind of ambiguity is accommodated in the way that, with appropriate data, the system may create two or more multiple alignments that have good scores. With {\em Fruit flies like a banana}, the two parsings can be seen in \cite[Figure 5.1]{wolff_2006}. Another example is the ambiguity in the phoneme sequence `\texttt{ae i s k r ee m}',\footnote{The symbols used are an alphabetic adaptation of phoneme symbols in the International Phonetic Alphabet.} which can be read as {\em ice cream} or {\em I scream}.

The way in which the SP system may use context to disambiguate `\texttt{ae i s k r ee m}' is described in \cite[Section 5.2.2]{wolff_2006} with examples---multiple alignments with phonetic versions of {\em I scream loudly} and {\em Ice cream is cold}---in \cite[Figure 5.4]{wolff_2006}.

\subsection{Integration of vision with other senses and other aspects of intelligence}\label{vision_integration_section}

It is clear that in people and other animals, vision does not stand alone but works in close association with other senses. Our concept of a ship, for example, is an amalgam of images, sounds, smells, the flavour of food on board, textures of different surfaces, and so on. In a similar way, natural vision works closely with other aspects of intelligence: unsupervised learning, several kinds of reasoning, understanding and producing natural language, recalling information, and non-visual kinds of recognition.

Achieving these kinds of integration without undue complexity has been a central aim in the development of the theory. And in that development, many candidate ideas have been rejected because they did not help to promote the simplification and integration of concepts. Now, the main planks of the theory are: representing all kinds of knowledge with patterns in one or two dimensions; the multiple alignment concept as it has been developed in the SP theory; and the overarching role of information compression via the matching and unification of patterns, in both the representation and processing of knowledge.

In artificial systems, the integration of vision with other sensory modalities and other aspects of intelligence is necessary for the development of versatility and adaptability in such systems.

\section{Conclusion}

As described in the Introduction, the main aim of this paper has been to explore how the SP theory may be applied to the understanding of natural vision and the development of computer vision, and to discuss associated issues.

The SP theory has things to say about several aspects of vision:

\begin{itemize}

\item Low level perceptual features such as edges or corners may be identified by the extraction of redundancy in uniform areas in a manner that is analogous to the run-length encoding technique for information compression, and comparable with the effect of lateral inhibition in the visual systems of animals.

\item The concept of {\em multiple alignment} in the SP theory may be applied to the recognition of objects, and to scene analysis, with a hierarchy of parts and sub-parts, and at multiple levels of abstraction.

\item The theory has potential for the unsupervised learning of visual objects and classes of objects, and suggests how coherent concepts may be derived from fragments. It provides an account of how we may discover objects via stereo matching and via motion.

\item As in natural vision, both recognition and learning in the SP system is robust in the face of errors of omission, commission and substitution.

\item The theory suggests how, via vision, we may piece together a knowledge of the three-dimensional structure of objects and of our environment that is good enough for many practical purposes.

\item The theory provides an account of how we may see things that are not objectively present in an image.

\item The theory suggests how we may recognise something despite variations in the size of its retinal image. It also suggests how images may be represented in a way that combines the advantages of raster graphics and vector graphics.

\item The theory has things to say about the phenomena of lightness constancy and colour constancy, about the role of context in recognition, and about ambiguities in visual perception.

\end{itemize}

A strength of the SP theory is that it is not simply a theory of vision. It is designed to achieve an overall simplification and integration of concepts in computing and cognition. There is clear potential for the integration of vision with other sensory modalities and with other aspects of intelligence such as unsupervised learning, processing of natural languages, reasoning, planning, and problem solving, with consequent benefits in terms of versatility and adaptability.

A high-parallel, open-source version of the SP machine, as outlined in \cite[Section 3.2]{sp_extended_overview}, would provide a means for researchers to explore what can be done with the system and to create new versions of it.

\section*{Competing interests}

The author has no competing interests.

\bibliographystyle{bmc-mathphys}


\end{document}